\documentclass[runningheads]{llncs}
\usepackage[T1]{fontenc}
\usepackage{graphicx}
\usepackage{amsfonts}
\usepackage{url}
\usepackage{xcolor}
\usepackage{soul}
\begin{document}
\title{YOLOv8-ResCBAM: YOLOv8 Based on\\
An Effective Attention Module for\\
Pediatric Wrist Fracture Detection}
\titlerunning{YOLOv8-ResCBAM}
%
\author{Rui-Yang Ju\inst{1} \and
Chun-Tse Chien\inst{2} \and
Jen-Shiun Chiang\inst{2}}
\authorrunning{Ju et al.}
%
\institute{Graduate Institute of Networking and Multimedia, National Taiwan University,\\ No. 1, Sec. 4, Roosevelt Rd., Taipei City 106319, Taiwan\\
\email{jryjry1094791442@gmail.com} \and
Department of Electrical and Computer Engineering, Tamkang University,\\ No.151, Yingzhuan Rd., Tamsui Dist., New Taipei City, 251301, Taiwan\\
\email{popper0927@hotmail.com; jsken.chiang@gmail.com}}
\maketitle
\begin{abstract}
Wrist trauma and even fractures occur frequently in daily life, particularly among children who account for a significant proportion of fracture cases.
Before performing surgery, surgeons often request patients to undergo X-ray imaging first, and prepare for the surgery based on the analysis of the X-ray images.
With the development of neural networks, You Only Look Once (YOLO) series models have been widely used in fracture detection for Computer-Assisted Diagnosis, where the YOLOv8 model has obtained the satisfactory results.
Applying the attention modules to neural networks is one of the effective methods to improve the model performance.
This paper proposes YOLOv8-ResCBAM, which incorporates Convolutional Block Attention Module integrated with resblock (ResCBAM) into the original YOLOv8 network architecture.
The experimental results on the GRAZPEDWRI-DX dataset demonstrate that the mean Average Precision calculated at Intersection over Union threshold of 0.5 (mAP 50) of the proposed model increased from 63.6\% of the original YOLOv8 model to 65.8\%, which achieves the state-of-the-art performance.
The implementation code is available at \url{https://github.com/RuiyangJu/Fracture_Detection_Improved_YOLOv8}.

\end{abstract}
\section{Introduction}
Wrist fractures are one of the most common fractures, particularly among the elderly and children~\cite{hedstrom2010epidemiology,randsborg2013fractures}, which mainly occur in the distal 2 cm of the radius near the joint.
Failure to provide timely treatment may result in deformities of the wrist joint, restricted joint motion, and joint pain for the patients~\cite{bamford2010qualitative}.
For children, a misdiagnosis would lead to a lifelong inconvenience~\cite{kraus2010treatment}.

In cases of pediatric wrist fractures, surgeons often inquire about the reasons leading to the fracture, and ask patients to conduct the fracture examination.
Fracture examinations are mainly conducted by three types of medical imaging equipment: Magnetic Resonance Imaging (MRI), Computed Tomography (CT), and X-ray.
Among them, X-ray is the preferred choice for most patients due to its cost-effectiveness~\cite{wolbarst1999looking}.
In hospitals providing advanced medical care, radiologists are required to follow the Health Level 7 (HL7) and Digital Imaging and Communications in Medicine (DICOM) international standards for the archival and transfer of X-ray images~\cite{boochever2004his}.
Nevertheless, the scarcity of radiologists in underdeveloped regions poses a challenge to the prompt delivery of patient care~\cite{burki2018shortfall,rimmer2017radiologist,rosman2015imaging}.
The studies in~\cite{erhan2013overlooked,mounts2011most} indicate a concerning 26\% error rate in medical imaging analysis during emergency cases.

Computer-Assisted Diagnosis (CAD) provides the experts (radiologists, surgeons, etc.) with help in some decisions.
With the development of deep learning~\cite{mahmud2021deep,mahmud2018applications} and the medical image processing techniques~\cite{adams2021artificial,choi2020using,chung2018automated,tanzi2020hierarchical}, more and more researchers are trying to employ neural networks for CAD, including fracture detection~\cite{bluthgen2020detection,gan2019artificial,kim2018artificial,lindsey2018deep,yahalomi2019detection}.

You Only Look Once (YOLO)~\cite{redmon2016you}, as one of the most important network models for object detection tasks~\cite{ju2024resolution}, shows the satisfactory model performance in fracture detection~\cite{hrvzic2022fracture}.
GRAZPEDWRI-DX~\cite{nagy2022pediatric} is a public dataset for fracture detection, which contains 20,327 pediatric wrist trauma X-ray images.
With the introduction of YOLOv8~\cite{jocher2023yolo}, it has been employed in this dataset for CAD~\cite{ju2023fracture}.

Due to the capacity of the attention modules to accurately focus on the important information of the input images, they are widely applied to neural network architectures.
Presently, there are two main types of attention modules: spatial attention and channel attention, designed to capture pixel-level pairwise relationships and channel dependencies, respectively~\cite{huang2019ccnet,lee2019srm,li2019expectation,zhao2018psanet,zhu2019asymmetric}.
Studies in~\cite{cao2019gcnet,fu2019dual,li2019selective} have demonstrated that incorporating attention module into convolutional blocks shows great potential for the model performance improvement.
Therefore, this paper proposes YOLOv8-ResCBAM, which integrates Convolutional Block Attention Module (CBAM)~\cite{woo2018cbam} integrates with resblock (ResCBAM) into the original YOLOv8 network architecture, to obtain the state-of-the-art (SOTA) on the GRAZPEDWRI-DX dataset~\cite{nagy2022pediatric}.

This paper's primary contributions are as follows:
\begin{itemize}
    \item It proposes YOLOv8-ResCBAM for pediatric wrist fracture detection by incorporating ResCBAM into the original YOLOv8 network architecture.
    \item It demonstrates that the proposed model significantly outperforms YOLOv8 model on the GRAZPEDWRI-DX dataset, achieving state-of-the-art performance.
    \item It employs YOLOv8-ResCBAM as a CAD tool to assist surgeons in analyzing X-ray images of wrist injuries.
\end{itemize}

This paper is organized as follows: Section~\ref{sec:related} introduces the researches on fracture detection utilizing deep learning methods, and outlines the evolution of attention modules.
Section~\ref{sec:method} presents the network architecture of the proposed model.
Section~\ref{sec:experiment} conducts a comparative experiment of the performance of the proposed model with the YOLOv8 (Baseline) model.
Finally, Section~\ref{sec:conclusion} concludes this research work, and discusses the future work.

\section{Related Work}
\label{sec:related}
\subsection{Fracture Detection}
Fracture detection is a hot topic in medical image analysis and processing, where the researchers usually employ neural networks on medical images, including the YOLO series models~\cite{bochkovskiy2020yolov4,jocher2023yolo,jocher2020yolov5,redmon2016you}.
Before the release of the GRAZPEDWRI-DX dataset by Nagy \emph{et al.}~\cite{nagy2022pediatric}, there were less publicly available pediatric datasets relevant to the pediatric wrist X-ray images.
Furthermore, the availability of musculoskeletal radiology collections for adult cases was also limited.
Hržić \emph{et al.}~\cite{hrvzic2022fracture} utilized the YOLOv4 \cite{bochkovskiy2020yolov4} model to detect fractures on the GRAZPEDWRI-DX dataset, which demonstrated that the YOLO series models can  improve the accuracy in diagnosing wrist injuries in children based on X-ray images.
Ju \emph{et al.}~\cite{ju2023fracture} created the ``Fracture Detection Using YOLOv8 App'' to aid surgeons in interpreting fractures on X-ray images, thereby reducing the risk of misdiagnosis and enhancing the information available for fracture surgery.

\subsection{Attention Module}
Squeeze-and-Excitation Network (SENet)~\cite{hu2018squeeze} initially proposed a mechanism to learn channel attention efficiently by applying Global Average Pooling (GAP) to each channel independently.
Subsequently, channel weights are generated using the Fully Connected layer and the Sigmoid function, leading to the good model performance.
Following the introduction of feature aggregation and feature re-calibration in SENet, some studies~\cite{chen20182,gao2019global} attempt to improve the SE block by capturing more sophisticated channel-wise dependencies.
Woo \emph{et al.}~\cite{woo2018cbam} combined the channel attention module with the spatial attention module, introducing CBAM to improve the representation capabilities of Convolutional Neural Networks (CNNs).
This attention module can be applied to different neural network architectures to improve the model performance.

\begin{figure}[t]
  \centering
  \includegraphics[width=\linewidth]{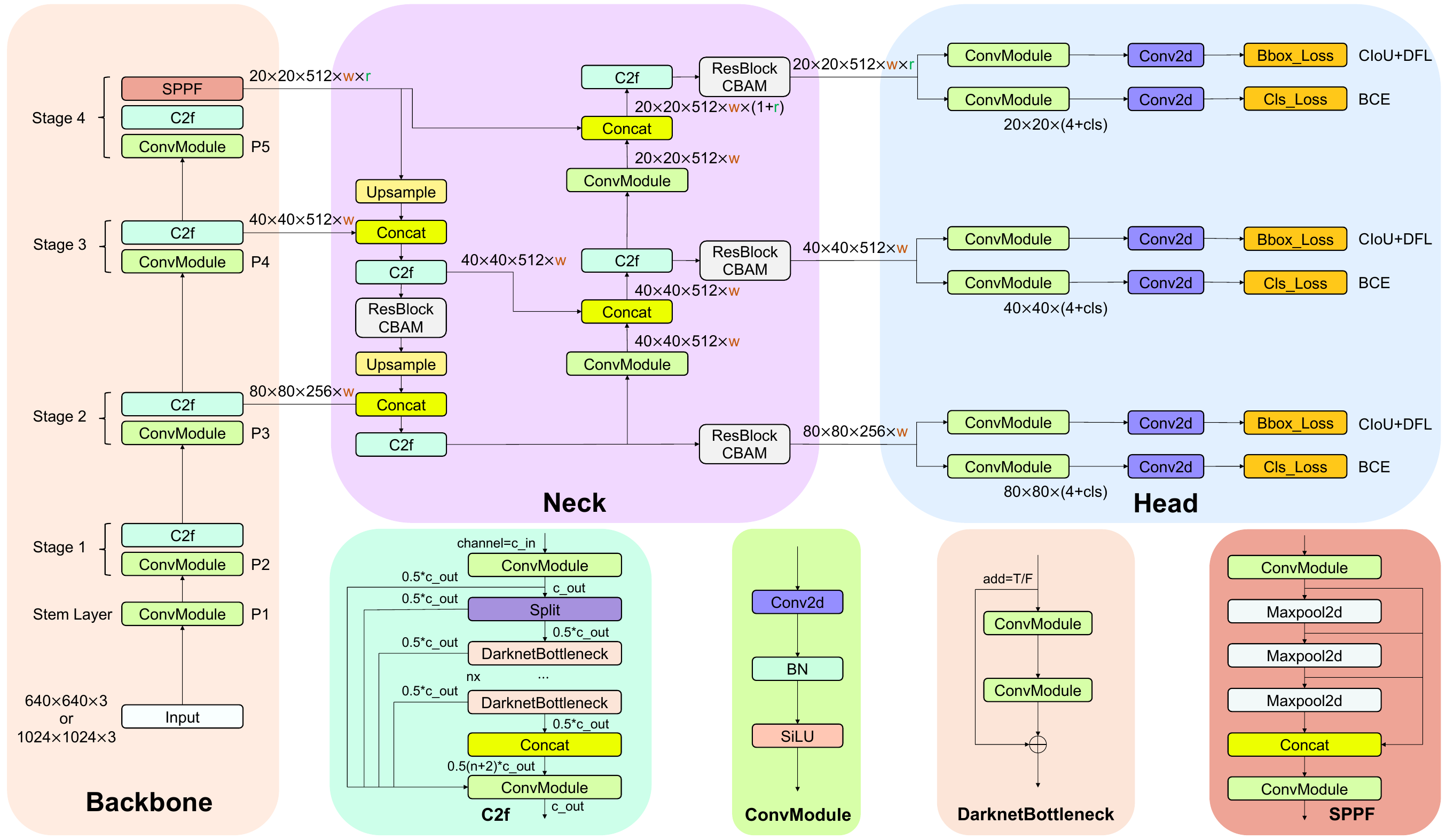}
  \caption{The network architecture of our proposed YOLOv8-ResCBAM model.}
  \label{fig_details}
\end{figure}

\section{Methodology}
\label{sec:method}
\subsection{Baseline}
YOLOv8 network architecture comprises four key components: Backbone, Neck, Head, and Loss Function.
The Backbone incorporates the Cross Stage Partial (CSP)~\cite{wang2020cspnet}, offering the advantage of reducing computational loads while enhancing the learning capability of CNNs.
As illustrated in Fig.~\ref{fig_details}, YOLOv8 differs from YOLOv5 employing the C3 module~\cite{jocher2020yolov5}, adopting the C2f module, which integrates the C3 module and the Extended ELAN (E-ELAN)~\cite{wang2022designing} from YOLOv7~\cite{wang2023yolov7}.
Specifically, the C3 module involves three convolutional modules and multiple bottlenecks, whereas the C2f module consists of two convolutional modules concatenated with multiple bottlenecks.
The convolutional module is structured as $Convolution - Batch Normalization - SiLU (CBS)$.

In the Neck part, YOLOv5 employs the Feature Pyramid Network (FPN)~\cite{lin2017feature} architecture for top-down sampling, ensuring that the lower feature map incorporates richer feature information.
Simultaneously, the Path Aggregation Network (PAN)~\cite{liu2018path} structure is applied for bottom-up sampling, enhancing the top feature map with more precise location information.
The combination of these two structures is executed to guarantee the accurate prediction of images across varying dimensions.
YOLOv8 follows the FPN and PAN frameworks while deleting the convolution operation during the up-sampling stage, as illustrated in Fig.~\ref{fig_details}.

In contrast to YOLOv5, which employs a coupled head, YOLOv8 adopts a decoupled head to separate the classification and detection heads.
Specifically, YOLOv8 eliminates the objectness branch, only retaining the classification and regression branches.
In addition, it departs from anchor-based~\cite{ren2015faster} method in favor of anchor-free~\cite{duan2019centernet} approach, where the location of the target is determined by its center, and the prediction involves estimating the distance from the center to the boundary.

In YOLOv8, the loss function employed for the classification branch involves the utilization of the Binary Cross-Entropy (BCE) Loss, as expressed by the equation as follows:
\begin{equation}
\label{eq:1}
Loss_{BCE} = -w[y_n \log_{}{x_n}+(1-y_n)\log_{}{(1-x_n)}],
\end{equation}
where $w$ denotes the weight; $y_n$ represents the labeled value, and $x_n$ signifies the predicted value generated by the model.

For the regression branch, YOLOv8 incorporated the use of Distribute Focal Loss (DFL)~\cite{li2020generalized} and Complete Intersection over Union (CIoU) Loss~\cite{zheng2021enhancing}.
The DFL function is designed to emphasize the expansion of probability values around object $y$, where the equation is presented as follows:
\begin{equation}
\label{eq:2}
Loss_{DFL} = -[(y_{n+1}-y)\log_{}{\frac{y_{n+1}-y_n}{y_{n+1}-y_n}}+(y-y_n)\log_{}{\frac{y-y_n}{y_{n+1}-y_n}}].
\end{equation}

The CIoU Loss introduces an influence factor to the Distance Intersection over Union (DIoU) Loss~\cite{zheng2020distance} by considering the aspect ratio of the predicted bounding box and the ground truth bounding box, where the corresponding equation is as follows:
\begin{equation}
\label{eq:3}
Loss_{CIoU} = 1-IoU+\frac{d^2}{c^2}+\frac{v^2}{(1-IoU)+v},
\end{equation}
where $IoU$ measures the overlap between the predicted and ground truth bounding boxes; $d$ is the Euclidean distance between the center points of the predicted and ground truth bounding boxes, and $c$ is the diagonal length of the smallest enclosing box that contains both predicted and ground truth bounding boxes.
In addition, $v$ represents the parameter quantifying the consistency of the aspect ratio, defined by the following equation:
\begin{equation}
\label{eq:4}
v = \frac{4}{\pi^2}(\arctan\frac{w_{gt}}{h_{gt}}-\arctan\frac{w_p}{h_p})^2,
\end{equation}
where $w$ denotes the weight of the bounding box; $h$ represents the height of the bounding box; $gt$ means the ground truth, and $p$ means the prediction.

\begin{figure}[t]
  \centering
  \includegraphics[width=\linewidth]{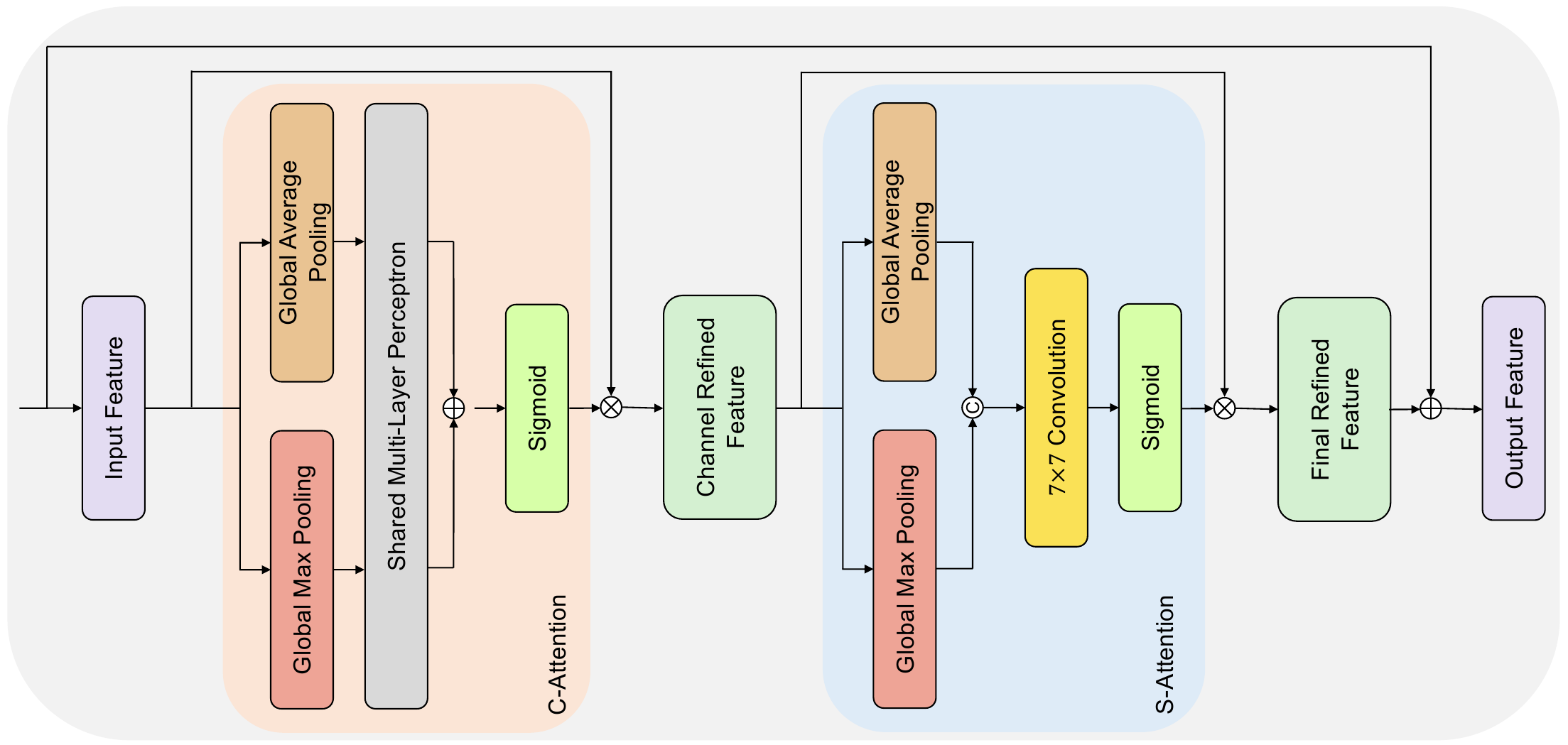}
  \caption{The network architecture of ResCBAM, which is Convolutional Block Attention Module (CBAM) integrated with resblock.}
  \label{fig_cbam}
\end{figure}

\subsection{Proposed Method}
In recent years, the attention modules have obtained excellent results in the field of object detection~\cite{jiang2022improved,li2020object,zhang2019object}.
With the integration of the attention modules, the networks can recognize the most important information of the input images for extraction and suppress the useless information.

This work incorporates an effective attention module into the Neck part of YOLOv8 to enhance the capture of key features and suppress the interfering information.
As illustrated in Fig.~\ref{fig_details}, we add a ResCBAM after each of the four C2f modules, where CBAM~\cite{woo2018cbam} has been shown to consistently improve classification and detection performance across various models.

CBAM comprises both channel attention (C-Attention) and spatial attention (S-Attention), as shown on Fig.~\ref{fig_cbam}.
Starting with an intermediate feature map $F_{input} \in \mathbb{R}_{C \times H \times W}$, CBAM first generates a 1D channel attention map $M_C \in \mathbb{R}^{C \times 1 \times 1}$, followed by a 2D spatial attention map $M_S \in \mathbb{R}^{1 \times H \times W}$, as formulated in the equation below:
\begin{equation}
\label{eq:5}
F_{CR} = M_C (F_{input}) \otimes F_{input},
\end{equation}

\begin{equation}
\label{eq:6}
F_{FR} = M_S (F_{CR}) \otimes F_{CR},
\end{equation}
where $\otimes$ is the element-wise multiplication; $F_{CR}$ is the Channel Refined Feature, and $F_{FR}$ is the Final Refined Feature.
For CBAM, $F_{output}$ is $F_{FR}$ as shown in the following equation:
\begin{equation}
\label{eq:7}
F_{output} = F_{FR}.
\end{equation}

It can be seen from Fig.~\ref{fig_cbam} that for ResCBAM, $F_{output}$ is the element-wise summation of $F_{input}$ and $F_{FR}$, as shown in the following equation:
\begin{equation}
\label{eq:8}
F_{output} = F_{input} + F_{FR}.
\end{equation}

Based on the previous studies~\cite{park2018bam,zeiler2014visualizing}, CBAM employs both Global Average Pooling (GAP) and Global Max Pooling (GMP) to aggregate the spatial information of a feature map, which generates two different spatial contextual descriptors.
Subsequently, these two descriptors share the same Multi-Layer Perceptron (MLP) with one hidden layer.
Finally, the output feature vectors from the element-wise summation are input to the sigmoid function ($\sigma$). The specific channel attention equation is as follows:
\begin{equation}
\label{eq:9}
M_C (F)= \sigma[MLP(GAP(F)) + MLP(GMP(F))].
\end{equation}

For the spatial attention, CBAM performs GAP and GMP along the channel axis respectively, and then concatenates them together to effectively highlight the information regions~\cite{zhou2016learning}, with the symbol $\copyright$ denoting concatenation.
Subsequently, a $7\times7$ convolutional layer is used to perform the convolution operation on these features.
The output of this convolution is used as the input of the sigmoid function ($\sigma$). The spatial attention is computed using the following equation:
\begin{equation}
\label{eq:10}
M_S (F)= \sigma[f^{7 \times 7}(GAP(F) \copyright GMP(F))].
\end{equation}

\section{Experiment}
\label{sec:experiment}
\subsection{Dataset}
GRAZPEDWRI-DX~\cite{nagy2022pediatric} is a public dataset of 20,327 pediatric wrist trauma X-ray images.
These X-ray images were collected by multiple pediatric radiologists in the Department for Pediatric Surgery of the University Hospital Graz between 2008 and 2018, involving 6,091 patients and a total of 10,643 studies.
This dataset is annotated with 74,459 image labels, featuring a total of 67,771 labeled objects.
This dataset exhibits a class imbalance, with 23,722 and 18,090 labels for ``text'' and ``fracture'', respectively, compared to only 276 and 464 labels for ``bone anomaly'' and ``soft tissue'', respectively.

\subsection{Preprocessing and Data Augmentation}
In the absence of predefined training, validation, and test sets by the publisher, we perform a random division, allocating 70\% to the training set, 20\% to the validation set, and 10\% to the test set.
Specifically, the training set comprises 14,204 images (69.88\%); the validation set includes 4,094 images (20.14\%); and the test set comprises 2,029 images (9.98\%).

Due to the limited diversity in brightness among the X-ray images within the training set, the model trained only on these images may not perform well in predicting other X-ray images.
Furthermore, the dataset suffers from class imbalance.
To address these issues, we employ data augmentation techniques to expand the training set.
Specifically, we fine-tune the contrast and brightness of the X-ray images using the addWeighted function available in the open-source computer vision library (OpenCV).

\subsection{Evaluation Metric}
In this study, we evaluate the performance of the YOLOv8-ResCBAM and YOLOv8 models in real-world diagnostic scenarios.
This work compares these models in terms of Parameters (Params), Floating Point Operations (FLOPs), F1 Score, the mean Average Precision calculated at Intersection over Union threshold of 0.5 (mAP 50), the mean Average Precision calculated at Intersection over Union thresholds from 0.5 to 0.95 (mAP 50-95), and inference time.

\begin{table}[ht]
\centering
\setlength{\tabcolsep}{1.2mm}{
\begin{tabular}{lccccccc}
\hline \noalign{\smallskip}
\textbf{Model} & \textbf{\begin{tabular}[c]{@{}c@{}}Input\\Size\end{tabular}} & \textbf{\begin{tabular}[c]{@{}c@{}}Params\\(M)\end{tabular}} & \textbf{\begin{tabular}[c]{@{}c@{}}FLOPs\\(B)\end{tabular}} & \textbf{\begin{tabular}[c]{@{}c@{}}F1\\Score\end{tabular}} & \textbf{\begin{tabular}[c]{@{}c@{}}mAP$\rm ^{val}$\\50\end{tabular}} & \textbf{\begin{tabular}[c]{@{}c@{}}mAP$\rm ^{val}$\\50-95\end{tabular}} & \textbf{\begin{tabular}[c]{@{}c@{}}Inference\\(ms)\end{tabular}} \\ \noalign{\smallskip} \hline \noalign{\smallskip} \hline \noalign{\smallskip}
Baseline-S & 640 & 11.13 & 28.5 & 0.59 & 60.4\% & 38.3\% & 1.9 \\
Baseline-M & 640 & 25.84 & 78.7 & 0.60 & 62.1\% & 38.9\% & 2.7 \\
Baseline-L & 640 & 43.61 & 164.9 & 0.61 & 62.4\% & 39.3\% & 3.9 \\ \noalign{\smallskip} \hline \noalign{\smallskip}
+ResCBAM-S & 640 & 16.06 & 38.3 & 0.62 & 61.6\% & 38.9\% & 1.9 \\
+ResCBAM-M & 640 & 33.84 & 98.2 & 0.62 & 62.8\% & 39.8\% & 2.9 \\
+ResCBAM-L & 640 & 53.87 & 196.2 & 0.63 & 62.9\% & 40.1\% & 4.1 \\ \noalign{\smallskip} \hline \noalign{\smallskip} \hline \noalign{\smallskip}
Baseline-S & 1024 & 11.13 & 28.5 & 0.60 & 62.5\% & 39.9\% & 2.8 \\
Baseline-M & 1024 & 25.84 & 78.7 & 0.61 & 62.6\% & 40.1\% & 5.2 \\
Baseline-L & 1024 & 43.61 & 164.9 & 0.62 & 63.6\% & 40.4\% & 7.8 \\ \noalign{\smallskip} \hline \noalign{\smallskip}
+ResCBAM-S & 1024 & 16.06 & 38.3 & 0.62 & 63.2\% & 39.0\% & 3.0 \\
+ResCBAM-M & 1024 & 33.84 & 98.2 & 0.63 & 64.3\% & 41.5\% & 5.7 \\
+ResCBAM-L & 1024 & 53.87 & 196.2 & 0.64 & 65.8\% & 42.2\% & 8.7 \\ \noalign{\smallskip} \hline \noalign{\smallskip}
\end{tabular}}
\caption{Results of the ablation study on the ResCBAM for fracture detection.}
\label{tab:ablation}
\end{table}

\subsection{Experiment Setup}
We train the YOLOv8 model and YOLOv8-ResCBAM models on the dataset~\cite{nagy2022pediatric}.
In contrast to the 300 epochs recommended by Ultralytics~\cite{jocher2023yolo} for YOLOv8 training, the experimental results of~\cite{ju2023fracture} indicate that the best performance is achieved within 60 to 70 epochs.
Consequently, this work sets 100 epochs for all models training.

For the hyperparameters of model training, we select the SGD optimizer instead of the Adam optimizer based on the result of the ablation study in~\cite{ju2023fracture}.
Following the recommendation of Ultralytics~\cite{jocher2023yolo}, this work establishes the weight decay of the optimizer at $5\times10^{-4}$, coupled with a momentum of 0.937, and the initial learning rate to $1\times10^{-2}$.
To compare the effects of different input image sizes on the performance of the models, this work sets the input image size to 640 and 1024 for the experiments respectively.

This work employs Python 3.9 for training all models on the framework of PyTorch 1.13.1. We advise readers to utilize versions higher than Python 3.7 and PyTorch 1.7 for model training, and the specific required environment can be accessed on our GitHub repository.
All experiments are executed using one single NVIDIA GeForce RTX 3090 GPU, with the batch size of 16 set to accommodate GPU memory constraints.

\subsection{Ablation Study}
In the ablation study, we investigate the impact of the ResCBAM on the model performance with input images of different sizes.
Specifically, we train our model using training sets with the input image sizes of 640 and 1024, respectively, and evaluate the YOLOv8-ResCBAM model on a test set with corresponding image sizes.
The results of the ablation study are presented in Table~\ref{tab:ablation}.
The F1 score and mAP 50 value achieved by the YOLOv8-ResCBAM model surpass those of the YOLOv8 (Baseline) model.
Specifically, when the input image size is 1024, the mAP 50 of the YOLOv8-ResCBAM-L model increases from 63.6\% to 65.8\% compared to the YOLOv8-L model, representing an improvement of 3.5\%.
In addition, the inference time of the YOLOv8-ResCBAM model (8.7ms/per image) is comparable to that of the YOLOv8 model (7.8ms/per image).

\begin{table}[ht]
\setlength{\tabcolsep}{1.5mm}{
\begin{tabular}{lcccccc}
\hline \noalign{\smallskip}
\textbf{Model} & \textbf{\begin{tabular}[c]{@{}c@{}}Params\\(M)\end{tabular}} & \textbf{\begin{tabular}[c]{@{}c@{}}FLOPs\\(B)\end{tabular}} & \textbf{\begin{tabular}[c]{@{}c@{}}F1\\Score\end{tabular}} & \textbf{\begin{tabular}[c]{@{}c@{}}mAP$\rm ^{val}$\\50\end{tabular}} & \textbf{\begin{tabular}[c]{@{}c@{}}mAP$\rm ^{val}$\\50-95\end{tabular}} & \textbf{\begin{tabular}[c]{@{}c@{}}Inference\\(ms)\end{tabular}} \\ \noalign{\smallskip} \hline \noalign{\smallskip} \hline \noalign{\smallskip}
YOLOv8 & 43.61 & 164.9 & 0.62 & 63.6\% & 40.4\% & 7.8 \\
YOLOv8-SA & 43.64 & 165.4 & 0.63 & 64.3\% & 41.6\% & 8.0 \\
YOLOv8-ECA & 43.64 & -65.5 & {\color{red}0.65} & 64.2\% & {\color{blue}41.9\%} & 7.7 \\
YOLOv8-MHSA & 44.69 & 166.9 & 0.63 & 63.7\% & 41.0\% & 15.6 \\
YOLOv8-GAM & 49.29 & 183.5 & {\color{red}0.65} & 64.2\% & 41.0\% & 12.7 \\
YOLOv8-ResGAM & 49.29 & 183.5 & {\color{blue}0.64} & {\color{blue}65.0\%} & 41.8\% & 18.1 \\
\textbf{Ours} & 53.87 & 196.2 & {\color{blue}0.64} & {\color{red}65.8\%} & {\color{red}42.2\%} & 8.7 \\ \noalign{\smallskip} \hline \noalign{\smallskip}
\end{tabular}}
\caption{Quantitative comparison of the performance of our YOLOv8-ResCBAM and other models on the GRAZPEDWRI-DX dataset when the input image size is 1024.}
\label{tab:comparison}
\end{table}

\begin{figure}[ht]
  \centering
  \includegraphics[width=\linewidth]{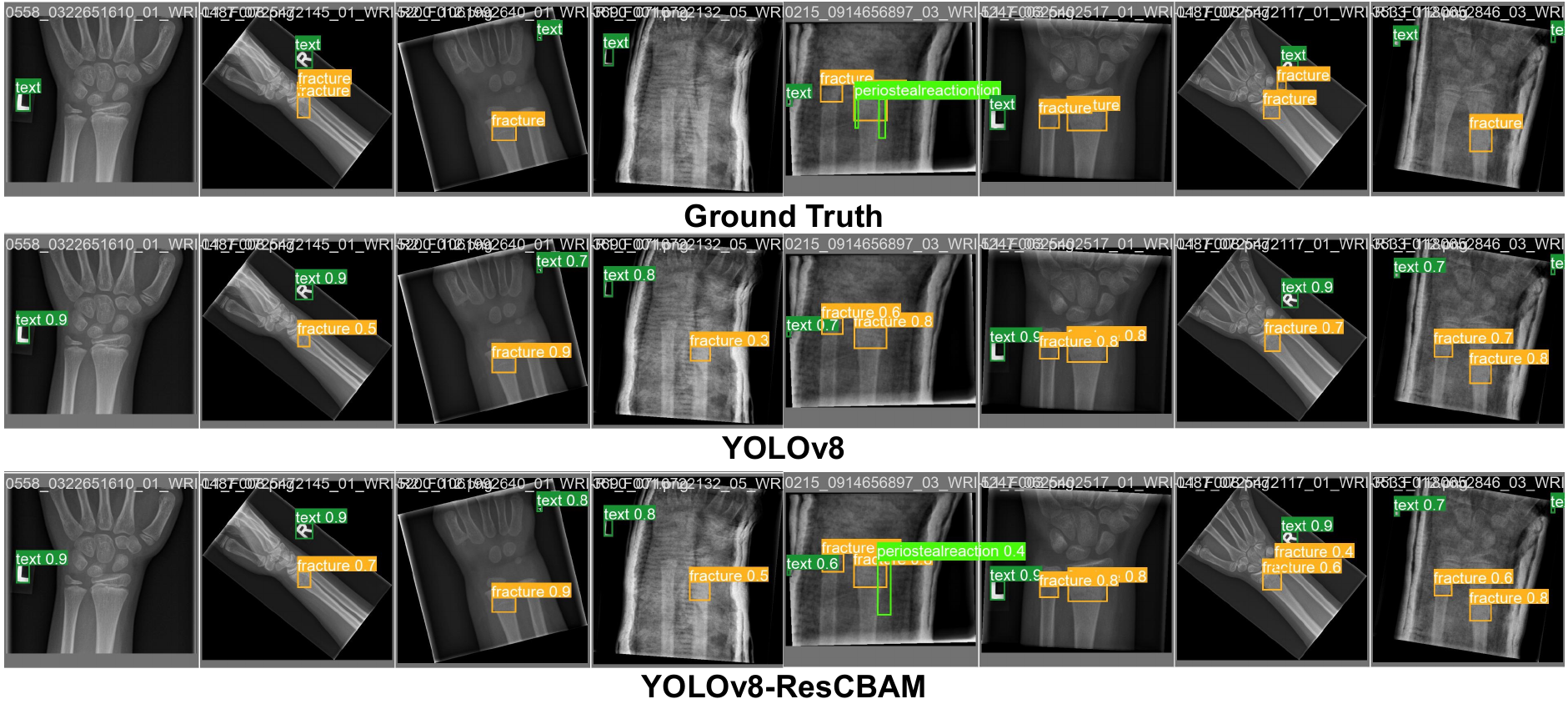}
  \caption{Examples of prediction results of our proposed model and YOLOv8 model for pediatric wrist fracture detection when the input image size is 1024.}
  \label{fig_prediction}
\end{figure}

\begin{figure}[ht]
  \centering
  \includegraphics[width=\linewidth]{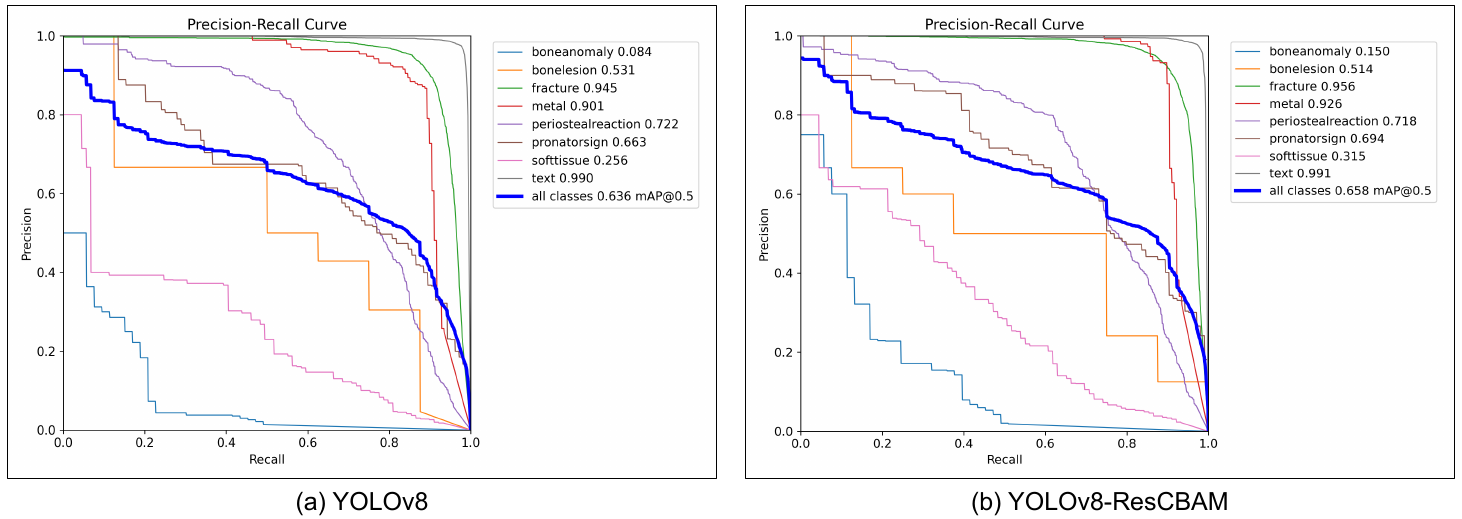}
  \caption{Visualization of the accuracy of predicting each class using our proposed and YOLOv8 models on the GRAZPEDWRI-DX dataset with the input image size of 1024.}
  \label{fig_pr}
\end{figure}

\subsection{Experimental Results}
As shown in Table~\ref{tab:ablation}, the performance of the models trained with the input image size of 1024 surpasses that of models trained using the same training set with the input image size of 640.
Nevertheless, it is noteworthy that this improvement in the model performance is accompanied by an increase in the inference time.
For instance, considering the YOLOv8-ResCBAM model with the large model size, the mAP 50-95 value attains 42.2\% for the input image size of 1024, which is 5.24\% higher than that of 40.1\% obtained for the input image size of 640.
However, the inference time of the model increases from 4.1ms to 8.7ms.

To demonstrate that our proposed model achieves the SOTA performance on the GRAZPEDWRI-DX dataset, we compare it with different models under the conditions of the input size of 1024 and large model size, as shown in Table~\ref{tab:comparison}.
Before we propose YOLOv8-ResCBAM, the models trained on this dataset had the highest mAP 50 value of 65.0\% and the highest mAP 50-95 value of 41.9\%.
In contrast, the mAP 50 and mAP 50-95 values of our YOLOv8-ResCBAM model are 65.8\% and 42.2\%, which outperform the current SOTA level.

This paper evaluates the impact of CBAM on the accuracy of the YOLOv8 model in predicting fractures in real case diagnosis scenarios. Eight X-ray images are randomly selected for evaluation.
Fig.~\ref{fig_prediction} illustrates the prediction results of both the YOLOv8-ResCBAM and the YOLOv8 models, which demonstrates our superior fracture detection ability compared to the YOLOv8 model, especially in cases of single fractures.
For instance, in the fourth X-ray image in Fig.~\ref{fig_prediction}, YOLOv8 predicts 30\% probability of a fracture, whereas the proposed model predicts 50\% probability, correctly recognizing the location as a fracture in the ground truth image.
Therefore, the YOLOv8-ResCBAM model can perform as a CAD tool, assisting radiologists and surgeons in making diagnoses.

Fig.~\ref{fig_pr} illustrates the Precision-Recall Curve (PRC) for each class predicted by the YOLOv8-ResCBAM model.
The analysis shows that both the YOLOv8-ResCBAM and YOLOv8 models can in correctly detect fractures, metals, and text, achieving an average accuracy of over 90\%.
However, their abilities of detecting the ``bone anomalies'' and ``soft tissue'' are notably weak, which significantly effects the mAP 50 value of models.
Specifically, the YOLOv8-ResCBAM model shows higher accuracies in predicting these two classes, achieving 15.0\% and 31.5\%, respectively, compared to 8.4\% and 25.6\% of YOLOv8.
We consider this is due to the small number of objects within these two classes in the used dataset.
As described in GRAZPEDWRI-DX~\cite{nagy2022pediatric}, the number of bone anomaly and soft tissue accounts for 0.41\% and 0.68\% of the total number of objects, respectively.
Consequently, any improvement in model performance via architectural enhancements is constrained by this data limitation.
To enhance the performance of the model, a recourse to incorporating extra data becomes imperative.

\section{Conclusion and Future Work}
\label{sec:conclusion}
Following the introduction of the YOLOv8, researchers began to employ it for fracture detection across various parts of the body.
Although the performance of the YOLOv8 model on the GRAZPEDWRI-DX dataset is commendable, it has not yet reached the SOTA level.
To address this limitation, we incorporate ResCBAM into the YOLOv8 architecture to enhance the model performances.
Notably, when the input image size is 1024, the mAP 50 of the YOLOv8-ResCBAM model obtains a superior performance of 65.8\%, surpassing the SOTA benchmark.
In addition our proposed YOLOv8-ResCBAM model can serve as a CAD tool aiding surgeons in the analysis of X-ray images, thereby reducing the possibility of misjudgment in fracture detection. 

In future work, we will try to collaborate with hospitals or medical colleges to collect more pediatric wrist X-ray images of unusual fractures or other diseases, including the labels ``bone anomaly'' and ``soft tissue'', to improve the model performance.
We hope that our proposed YOLOv8-ResCBAM model will be able to help experts (radiologists, surgeons, etc.) make more good decisions in the future.
Specifically, in our previous work~\cite{ju2023fracture}, we introduced the ``Fracture Detection Using YOLOv8 App'', and we plan to integrate the proposed YOLOv8-ResCBAM model into this application.

\subsubsection{Acknowledgment}
This work is supported by National Science and Technology Council of Taiwan, under Grant Number: NSTC 112-2221-E-032-037-MY2.

\bibliographystyle{splncs04}
\bibliography{5750}
\end{document}